\documentclass[preprint,12pt]{elsarticle}

\usepackage{amssymb}
\usepackage{amsmath}
\usepackage{multirow}
\usepackage{threeparttable}
\usepackage{overpic}
\usepackage[table]{xcolor}
\definecolor{myBlue}{HTML}{0000FF}
\definecolor{myPink}{HTML}{FF69B4}
\definecolor{myCyan}{HTML}{00B3B3}

\journal{Mechanism and Machine Theory}

\begin{document}
\begin{frontmatter}


\title{On Kinodynamic Global Planning in a Simplicial Complex Environment: A Mixed Integer Approach}

\author[1]{Otobong Jerome\corref{cor1}}
\ead{o.jerome@innopolis.university}
\author[2]{Alexandr Klimchik}
\ead{aklimchik@lincoln.ac.uk}
\author[1]{Alexander Maloletov}
\ead{a.maloletov@innopolis.university}
\author[2]{Geesara Kulathunga}
\ead{gkulathunga@lincoln.ac.uk}

\affiliation[1]{organization={Center of Autonomous Technologies, Innopolis University},
            addressline={Universitetskaya St, 1},
            city={Innopolis},
            postcode={420500},
            state={Republic of Tatarstan},
            country={Russia}}

\affiliation[2]{organization={School of Computer Science, University of Lincoln},
            addressline={Brayford Pool},
            city={Lincoln},
            postcode={LN6 7TS},
            state={Lincoln},
            country={United Kingdom}}

\cortext[cor1]{Corresponding author}

\begin{abstract}
This work casts the kinodynamic planning problem for car-like vehicles as an optimization task to compute a minimum-time trajectory and its associated velocity profile, subject to boundary conditions on velocity, acceleration, and steering. The approach simultaneously optimizes both the spatial path and the sequence of acceleration and steering controls, ensuring continuous motion from a specified initial position and velocity to a target end position and velocity.The method analyzes the admissible control space and terrain to avoid local minima.
The proposed method operates efficiently in simplicial complex environments, a preferred terrain representation for capturing intricate 3D landscapes. The problem is initially posed as a mixed-integer fractional program with quadratic constraints, which is then reformulated into a mixed-integer bilinear objective through a variable transformation and subsequently relaxed to a mixed-integer linear program using McCormick envelopes. Comparative simulations against planners such as MPPI and log-MPPI demonstrate that the proposed approach generates solutions $10^4$ times faster while strictly adhering to the specified constraints.
\end{abstract}




\end{frontmatter}



\section{Introduction}
It is becoming increasingly common to represent terrains in navigation systems using simplicial complexes, particularly 3D triangular meshes. These meshes effectively capture the 2D manifold nature of terrains, meaning that, while terrains exist in three-dimensional space, they locally resemble a continuous 2D surface. Even though the planned path consists of 3D points and must account for elevation changes, the planning itself is typically performed in a local 2D frame, considering vehicle dynamics such as acceleration and steering angle. Consequently, several mesh-based planning algorithms have been proposed \cite{cvp_putz, keenan2020}.

Unlike classical planning algorithms, which focus solely on geometric considerations, kinodynamic planning also accounts for constraints such as velocity and acceleration, adding significant complexity to the problem \cite{kinodynamicrrt, Webb2012KinodynamicRO}.

Deterministic methods like A* and Dijkstra’s algorithm guarantee completeness and optimality but are computationally expensive, especially in high-dimensional spaces. Probabilistic approaches, such as RRT and PRM, efficiently explore complex environments but struggle with optimality and tight constraints. While deterministic planners always find a solution if one exists, their high computational cost limits real-time use. In contrast, probabilistic planners generate trajectories quickly but often require multiple refinements to approach optimality \cite{zhao2022hybrid, likhachev2003ara, dolgov2010path, zheng2021accelerating}.

Additionally, in kinodynamic planning, \cite{tobiasKunz} demonstrated that the most commonly used RRT method of selecting the optimal input with a fixed time step is not probabilistically complete.

Furthermore, data-driven approaches, employing machine learning techniques such as neural networks and reinforcement learning, have been leveraged to identify patterns within data and predict viable trajectories. While flexible, they often struggle with generalisation and constraint compliance \cite{10023708,10238717,10011923,8460798}.  
In contrast, optimisation-based planners, such as those using Mixed-Integer Programming, explicitly handle constraints and can provide globally optimal solutions. These methods excel at integrating discrete decisions with continuous trajectory optimisation, but their performance depends on the specific problem formulation.\cite{abichandani2013mathematical, huang2021efficient,ding2009optimal}.

Existing local planning methods, such as Model Predictive Path Integral (MPPI), Model Predictive Control (MPC), and Iterative Linear Quadratic Regulator (iLQR), prioritise rapid and efficient responses to dynamic environments by focusing on immediate surroundings. While these approaches enable quick adaptation to local changes, they often overlook broader contextual information, which can lead to suboptimal long-term solutions. Conversely, global planning methods incorporate the entirety of the environment to optimise trajectories, examining overarching objectives, yielding globally optimal paths. However, this comprehensive approach is computationally intensive and less responsive to real-time environmental variations.

Although extensive research exists on local planning, 2D environments, and purely geometric settings, the 3D mesh domain, which requires integrating kinodynamic constraints, still lacks a deterministic global kinodynamic planner for car-like vehicles.

This paper presents the problem of kinodynamic planning for car-like vehicles on a 3D mesh as finding the shortest-time trajectory from a starting point to an endpoint on the mesh. The trajectory must follow a velocity profile that respects bounded constraints determined by the vehicle's physical limitations. This work proposes a Mixed-Integer Kinodynamic (MIKD) planner for car-like vehicles navigating a 3D mesh. This approach utilises modern solvers to optimise the planned path while adhering to the vehicle's physical constraints.

The main contributions of this work are as follows:
\begin{itemize}
    \item A  mixed integer formulation of the kinodynamic planning problem for car-like vehicles on terrains modelled using 3D meshes.
    \item An efficient algorithm to solve this mixed-integer kinodynamic planning problem.
    \item An open-source implementation of the global planner software.
\end{itemize}

In the following sections, we will review related works, establish the theoretical foundations, describe the algorithmic implementation, and present experimental results, highlighting the potential of our mixed-integer approach for kinodynamic planning in simplicial complex environments.

\section{Related Work}
Kinodynamic planning, which addresses both kinematic constraints and dynamic feasibility, has been an active research area since the seminal work of \cite{L1,L2}, where the problem was finding the minimal time trajectories subject to velocity and acceleration constraints. Since then, several applications of kinodynamic planning in robotics and vehicle motion planning have been explored \cite{kulathunga2021path,randomizedKinodynamic,gakd,kinodynamicrrt,bordalba2018randomized, kazemi2013randomized}. The solution to kinodynamic planning typically yields a map from time to generalised forces, as introduced in \cite{L1}.

\begin{table}[htbp]
    \centering
    \caption{A taxonomy of planners for autonomous navigation. Cells with a blue background indicate deterministic methods; those with a red background indicate probabilistic methods.}
    \label{tab:taxonomy}
    \renewcommand{\arraystretch}{1.3} 
    \begin{tabular}{c|c|p{3cm}}  
        \hline
        \textbf{Planner Type} & \textbf{Category} & \textbf{Methods} \\
        \hline
        \multirow{6}{*}{Global Planners} 
        & \cellcolor{blue!20}Optimization & \cellcolor{blue!20}Mixed-Integer \\ \cline{2-3}
        & \cellcolor{blue!20}Graph-Based & \cellcolor{blue!20}A*, Dijkstra, D*, Wavefront \\ \cline{2-3}
        & \cellcolor{blue!20}Decision-Theoretic & \cellcolor{blue!20}Markov Decision Processes (MDP) \\ \cline{2-3}
        & \cellcolor{red!20}Sampling-Based & \cellcolor{red!20}RRT, PRM \\ \cline{2-3}
        & \cellcolor{red!20}Learning-Based & \cellcolor{red!20} Imitation Learning, RL \\ 
        \hline
        \multirow{4}{*}{Local Planners} 
        & \cellcolor{blue!20}Optimization & \cellcolor{blue!20}MPC, NMPC, iLQR \\ \cline{2-3}
        & \cellcolor{red!20}Bayesian Optimization & \cellcolor{red!20}Gaussian Process Regressor \\ \cline{2-3}
        & \cellcolor{red!20}Learning-Based& \cellcolor{red!20}Deep Neural Networks, LSTM \\ \cline{2-3}
        & \cellcolor{red!20}Random Sampling & \cellcolor{red!20}MPPI \\ \cline{2-3}
        \hline
    \end{tabular}
\end{table} 

As summarized in Table \ref{tab:taxonomy}, in the local planning setting, popular methods include Iterative Linear Quadratic Regulator (ILQR)\cite{zeng2023i2lqr}, Model Predictive Path Integral (MPPI)\cite{williams2017model,buyval2019model}, Trajectory Optimization with sequential convex optimization (TrajOpt) \cite{schulman2013finding}, \cite{matiussi2018trajectory} and Covariant Hamiltonian Optimization for Motion Planning (CHOMP) \cite{ratliff2009chomp}. While effective in generating feasible trajectories, these methods are inherently limited to short planning horizons and lack a global understanding of the terrain, necessitating approaches like Global-MPPI \cite{LI2024109645}.

Efforts to integrate kinodynamic constraints into global planning algorithms like Rapidly-exploring Random Trees (RRT) and A* have shown promise but face challenges, especially when applied to a 3D triangular mesh\cite{zhao2022hybrid, likhachev2003ara,dolgov2010path,zheng2021accelerating,webb2013kinodynamic}. RRT excels in exploring high-dimensional spaces but often lacks optimality and consistency in trajectory smoothness \cite{zhao2022hybrid}. A*, on the other hand, guarantees optimality under heuristic-guided search but becomes computationally intractable when dealing with high-dimensional constraints and non-convex cost functions. Data-driven approaches such as \cite{10023708,10238717,10011923,8460798}, using machine learning and deep reinforcement learning, are common, although they often struggle with generalisation and constraint compliance. 
\begin{table}[htbp]
\centering
\caption{Summary of Mixed-Integer Programming in Motion Planning Literature}
\label{tab:lit_summary}
\resizebox{\linewidth}{!}{%
\begin{tabular}{l|l|l|l|l}
\hline
\textbf{Works} & \textbf{Terrain} & \textbf{Application} & \textbf{MIP Type} & \textbf{Key Constraints} \\
\hline
Reiter et al., 2025 \cite{r10} & 2D & Multi-lane Traffic & MIQP & Collision avoidance, acceleration, safety margins \\
Quirynen et al., 2025 \cite{r11} & 2D & Lane change & MIQP & Lane change, collision avoidance, real-time feasibility \\
Caregnato-Neto \& Ferreira, 2025 \cite{r18} & 2D & Micro-mobility vehicle & MILP & Intersample collision avoidance, orientation \\
Robbins et al., 2024 \cite{r1} & 2D & Autonomous vehicles & MIQP & Obstacle avoidance, state/input, risk-aware cost \\
Jaitly \& Farzan, 2024 \cite{r4} & -- & Pendulum swing-up & MILP & Torque limits \\
Gratzer et al., 2024 \cite{r8} & 2D & Connected vehicles & MI-MPC & Obstacle avoidance, soft state constraints \\
Caregnato-Neto et al., 2024 \cite{r20} & 2D & Nonholonomic robots & MILP & VLC connectivity, nonholonomic constraints \\
Bhattacharyya \& Vahidi, 2023 \cite{r17} & 2D & Highway merging & MI-MPC & Vehicle costs, adaptive cost function \\
Battagello et al., 2021 \cite{r9} & 2D & Mobile robots & MILP & Collision avoidance, reduced binary variables \\
Ding et al., 2020 \cite{r6} & -- & Two-legged robot & MICP & Centroidal motion, contact, wrench, torque, friction \\
Esterle et al., 2020 \cite{r16} & 2D & Autonomous driving & MIQP & Nonholonomic motion, steering, acceleration limits \\
Ding et al., 2018 \cite{r7} & -- & Single-leg robots & MIQCP & Actuator torque, workspace polytopes, rough terrain \\
Dollar \& Vahidi, 2018 \cite{r15} & 2D & Lane switching & MI-MPC & Longitudinal control, lane switching, fuel efficiency \\
Gawron \& Michałek, 2018 \cite{r22} & 2D & Unicycle & MILP & Bounded curvature, waypoint orientation \\
Altché et al., 2016 \cite{r21} & 2D & Mobile robots & MILP & Kinodynamic constraints, time discretization \\
Deits \& Tedrake, 2014 \cite{r5} & -- & Humanoid robots & MIQCQP & Kinematic reachability, rotation, obstacle avoidance \\
Ding et al., 2011 \cite{r2} & -- & Robotic manipulators & MILP & Polyhedral obstacles, joint velocity, kinematic/dynamic \\
Shengxiang \& Pei, 2008 \cite{r12} & -- & UAV & MILP & Velocity/acceleration, terrain avoidance \\
Ma \& Miller, 2006 \cite{r13} & 3D & Local Planning & MI-MPC & Obstacle avoidance, velocity, terrain modeling \\
Richards \& How, 2005 \cite{r14} & -- & Local Planning & MI-MPC & Stability, feasibility, robustness \\
\cellcolor{blue!20} Proposed & \cellcolor{blue!20} 3D & \cellcolor{blue!20} Global Planning & \cellcolor{blue!20} MILP & \cellcolor{blue!20} Kinodynamics, terrain modeling, collision avoidance \\
\hline
\end{tabular}
}
\end{table}
Mixed-integer programming (MIP) is a well-established tool in motion planning, valued for its ability to model continuous dynamics and discrete decisions. This structured approach effectively handles complex kinodynamic constraints, obstacle avoidance, and logical conditions. The work \cite{r19} thoroughly reviews the subject. 

The literature employs various MIP formulations, each tailored to specific motion planning challenges as summarised in Table \ref{tab:lit_summary}, most commonly for local planning, where it is used within the receding horizon framework to optimise over a finite horizon. 

Mixed-Integer Linear Programming (MILP) is notable for its computational efficiency. It has been effectively employed in robotic manipulator path planning, where geometric techniques reduced binary variables \cite{r2}, and in UAV trajectory planning by converting nonlinear terrain avoidance constraints into linear inequalities \cite{r12}. For non-holonomic robots and micro-mobility vehicles, MILP enabled intersample collision avoidance and addressed visible light communication (VLC) constraints \cite{r18, r20}, and for torque-constrained pendulum swing-up problems with polytopic action sets \cite{r4}. MILP excels in problems with linear dynamics and constraints but may struggle with non-linearities unless approximated.

Mixed-Integer Quadratic Programming (MIQP) extends MILP by incorporating quadratic objectives, making it well-suited for modelling energy and smoothness costs. MIQP has been applied in autonomous vehicle motion planning using hybrid zonotopes and convex relaxations for obstacle avoidance \cite{r1}, and in real-time decision-making and lane switching with custom solvers \cite{r11, r15}. MIQP has also addressed convex sub-polygon constraints in autonomous driving \cite{r16}, offering a balance between computational efficiency and expressive modelling, though it demands careful relaxation for non-convexities.

Mixed-Integer Nonlinear Programming (MINLP) handles nonlinear objectives and constraints, offering flexibility for complex dynamics. It was used for multi-robot coordination, later refined to MILP for computational efficiency \cite{r3}. MINLP is computationally intensive but suitable for problems with nonlinear kinodynamic constraints, though it may sacrifice real-time feasibility.
Mixed-Integer Quadratically Constrained Quadratic Programming (MIQCQP) incorporates quadratic constraints, which are ideal for problems with complex geometric or dynamic restrictions. It was applied for humanoid footstep planning, using convex inner approximations for kinematic reachability \cite{r5}. It was also used for single-leg dynamic motion planning, addressing actuator torque and rough terrain \cite{r7}. MIQCQP is computationally demanding but effective for problems requiring precise modelling of quadratic constraints.

Mixed-Integer Convex Programming (MICP) deals with convex constraints, balancing expressiveness and solvability. MICP was used in \cite{r6}  for multi-legged robot jumping, incorporating piecewise convexification for dynamics. MICP is less computationally intensive than MINLP but requires convex approximations, limiting its applicability to highly nonlinear systems.

Recent advancements include hybrid architectures to improve real-time performance. A two-layer model predictive control (MPC) architecture was proposed for connected automated vehicles, combining an upper-level MIQP for global optimality with a lower-level quadratic programming (QP) MPC for real-time collision avoidance using Big-M constraints \cite{r8}. Similarly, an equivariant deep learning approach is used to predict integer variables in MIQPs, thereby enhancing real-time decision-making for traffic scenarios \cite{r10}. These hybrid methods reduce computational overhead while maintaining solution quality.

The constraints in MIP-based motion planning vary depending on the application and formulation. Non-convex obstacle avoidance constraints in MIP-based motion planning are solved using Big-M techniques or convex relaxations. Convex hull relaxations were applied for autonomous vehicles in \cite{r1}, and  Big-M and linear half-space constraints in \cite{r8}. Dynamic obstacle clustering reduced binary variables for efficiency \cite{r9}. These methods balance between computational complexity and modelling accuracy.
Velocity, acceleration, and jerk constraints ensure that trajectories are physically feasible. Kinodynamic constraints for multi-robot coordination \cite{r3, r21}, and region-dependent acceleration and jerk limits in \cite{r16}. Velocity and acceleration limits were used for Unmanned Aerial Vehicles (UAVs) and real-time planning \cite{r12, r13}, increasing problem complexity while ensuring dynamic feasibility.
Terrain modelling, such as triangulated irregular networks (TIN), is used for UAVs and rough terrain navigation \cite{r12, r13}, and in  \cite{r5, r7}, polytopic constraints and rough terrain constraints are handled for legged robots.
Torque constraints for pendulum swing-up \cite{r4} and legged robots with torque and friction limits \cite{r6, r7} are modelled. These ensure actuator feasibility but increase optimisation complexity. Logical constraints, like waypoint selection, are modelled using binary variables \cite{r18}.

However, optimisation-based approaches often require a precise problem formulation tailored to the specific domain. This work presents a formulation for the motion of a car-like vehicle on a 3D mesh, considering the primary control constraints: the acceleration range, which directly affects the velocity profile, and the steering angle, which influences the path's curvature. The vehicle then tracks the fastest feasible path while respecting its steering angle and acceleration limits.

This work differs from existing mixed-integer formulations in several key aspects: the use of 3D meshes allowing for arbitrary terrain representation which is especially useful for off-road navigation, focus on vehicle control limitations rather than exact vehicle-terrain dynamic models, incorporation into the optimisation problem of a transition function which allows computation and elimination of non-traversable terrain regions, and a relaxation strategy that enables faster computation compared to MIQP approaches. Most of the existing literature focuses on a flat 2D terrain. Works such as \cite{r12, r13} use an uneven terrain representation, but \cite{r12} is specifically designed for UAVs, and \cite{r13} employs the receding horizon paradigm, which is prone to getting stuck in local minima rather than being efficient for global planning. 

The proposed approach supports a hierarchical implementation. An initial optimisation model, incorporating terrain-specific constraints, can be developed offline. Subsequently, vehicle-specific constraints, such as steering angle and acceleration, can be included. Finally, start and destination constraints can be applied online before solving the model.

This approach shifts the focus away from precise dynamic modelling of the vehicle, as required by existing gradient-based methods. Instead, it prioritises ensuring the generated path remains within the vehicle's traversable range. Also, a well-designed velocity profile minimises energy consumption by avoiding unnecessary speeding up and slowing down.

\section{Methodology}
The objective is to compute a feasible trajectory for a car-like vehicle that satisfies both geometric and kinodynamic constraints. This objective requires determining a continuous path that respects the vehicle's acceleration and steering angle limitations.

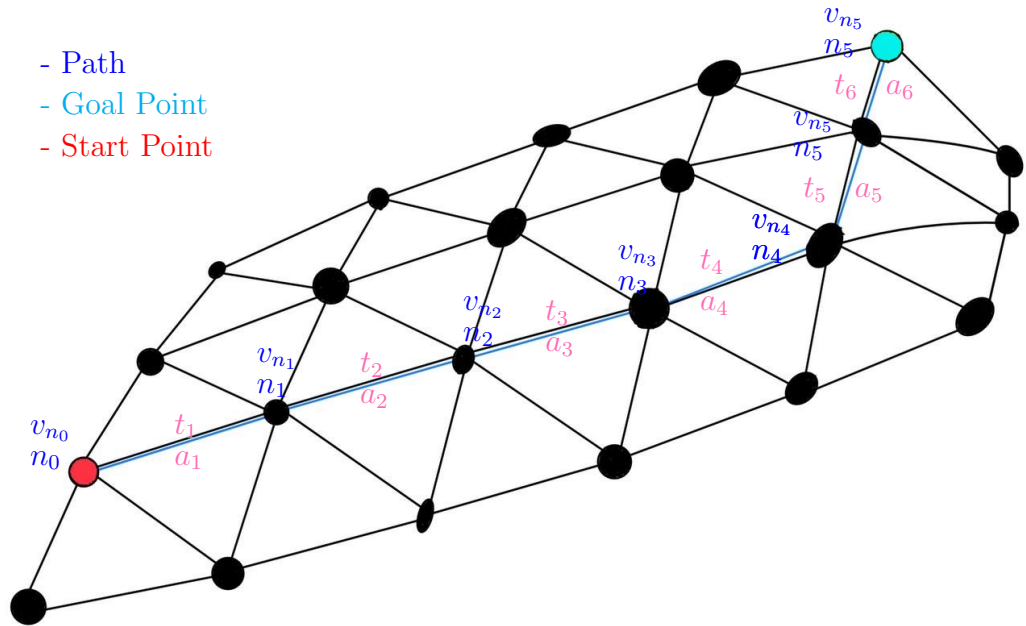
\begin{figure}[htp]
  \centering
  \begin{overpic}[width=\linewidth]{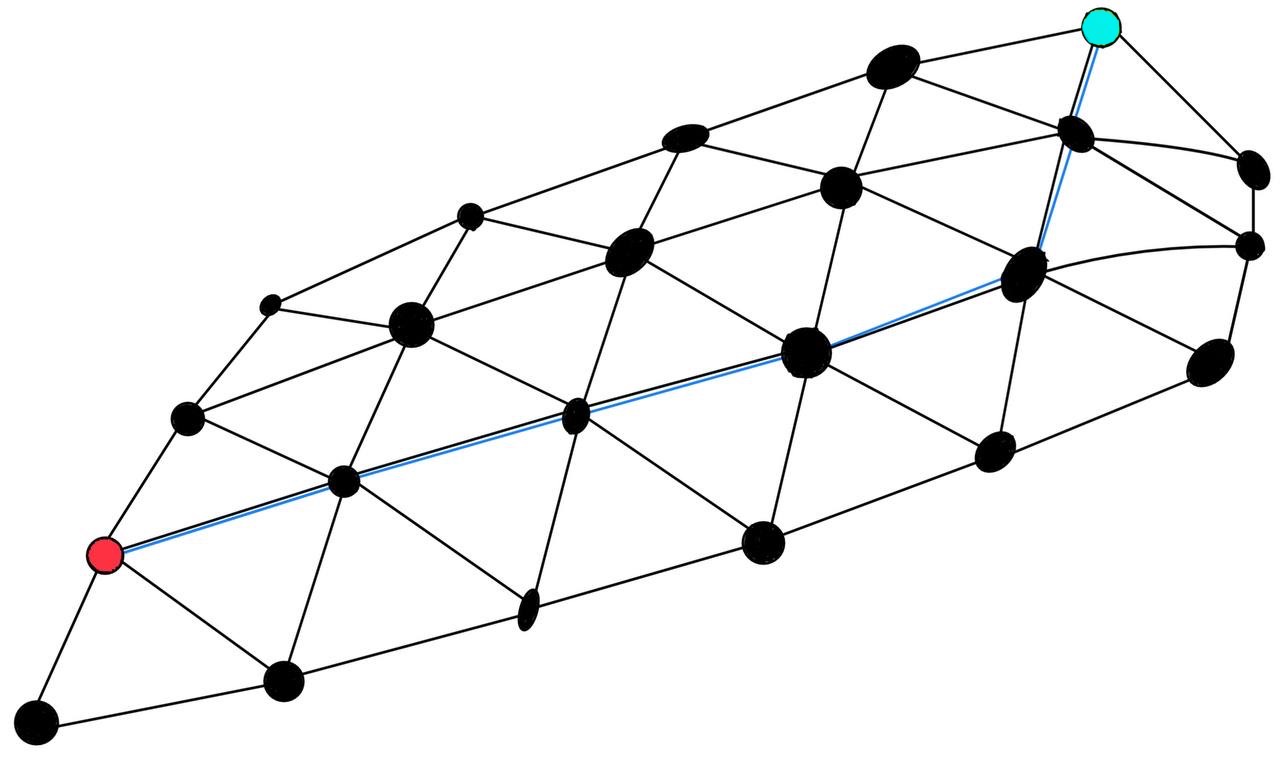}
    \put(4,47){\color{red}- Start Point}
    \put(4,51){\color{cyan}- Goal Point}
    \put(4,55){\color{myBlue} - Path}
    \put(3,20){\color{myBlue} $v_{n_0}$}
    \put(3,17){\color{myBlue} $n_0$}

    \put(17,20){\color{myPink} $t_{1}$}
    \put(17,17){\color{myPink} $a_1$}

    \put(25,27){\color{myBlue} $v_{n_1}$}
    \put(25,24){\color{myBlue} $n_1$}

    \put(35,26){\color{myPink} $t_{2}$}
    \put(35,23){\color{myPink} $a_2$}

    \put(45,32){\color{myBlue} $v_{n_2}$}
    \put(45,29){\color{myBlue} $n_2$}

    \put(53,31){\color{myPink} $t_{3}$}
    \put(53,28){\color{myPink} $a_3$}
    
    \put(60,37){\color{myBlue} $v_{n_3}$}
    \put(60,34){\color{myBlue} $n_3$}

    \put(68,36){\color{myPink} $t_{4}$}
    \put(68,32){\color{myPink} $a_4$}

    \put(73,40){\color{myBlue} $v_{n_4}$}
    \put(73,37){\color{myBlue} $n_4$}

    \put(78,43){\color{myPink} $t_{5}$}
    \put(83,43){\color{myPink} $a_5$}
    
    \put(73,40){\color{myBlue} $v_{n_4}$}
    \put(73,37){\color{myBlue} $n_4$}
    
    \put(77,50){\color{myBlue} $v_{n_5}$}
    \put(77,47){\color{myBlue} $n_5$}

    \put(81,53){\color{myPink} $t_{6}$}
    \put(86,53){\color{myPink} $a_6$}
    
    \put(80,60){\color{myBlue} $v_{n_5}$}
    \put(80,57){\color{myBlue} $n_5$}
  \end{overpic}
  \caption{Vehicle path through nodes from start to goal, with edge accelerations and times, meeting initial and final velocities, and max acceleration/curvature limits. A vehicle at node \( n_i \) with velocity \( v_{n_i} \), accelerating at \( a_i \) along the edge \( e_{n_i, n_{i+1}} \), for time \(t_i\) reaches node \( n_{i+1} \) with velocity \( v_{n_{i+1}} \). The objective is to determine a sequence of nodes that minimizes the total traversal time \( \sum t_i \), while ensuring that acceleration does not exceed \( a_{\max} \) and the curvature between consecutive edges remains within the vehicle's steering limit.}
    \label{fig:method}
\end{figure}
As shown in Figure~\ref{fig:method}, the problem is formulated in a discretised mesh environment, where the vehicle follows a sequence of nodes \( n_i \) connected by edges \( e_{n_i, n_{i+1}} \). At each node \( n_i \), the vehicle has a velocity \( v_{n_i} \) and accelerates at \( a_i \) for time \(t_i\) along the edge to reach the next node \( n_{i+1} \) with velocity \( v_{n_{i+1}} \). Since for every path segment \( e_{n_i, n_{i+1}} \), there exist valid acceleration and steering controls that allow the vehicle to transition from \( n_i \) to \( n_{i+1} \), it is possible to generate a trajectory from the start point with initial velocity \( v_{\text{init}} \) to the goal point with final velocity \( v_{\text{final}} \), without violating the acceleration and steering angle control limits. The goal is thus to determine a sequence of waypoints and velocities that minimises the total traversal time \( \sum t_i \), while ensuring compliance with acceleration and steering constraints. 

\subsection{Problem Formulation}
\begin{figure*}[htp]
  \centering
  \vspace{1cm}
  \begin{overpic}[width=\textwidth]{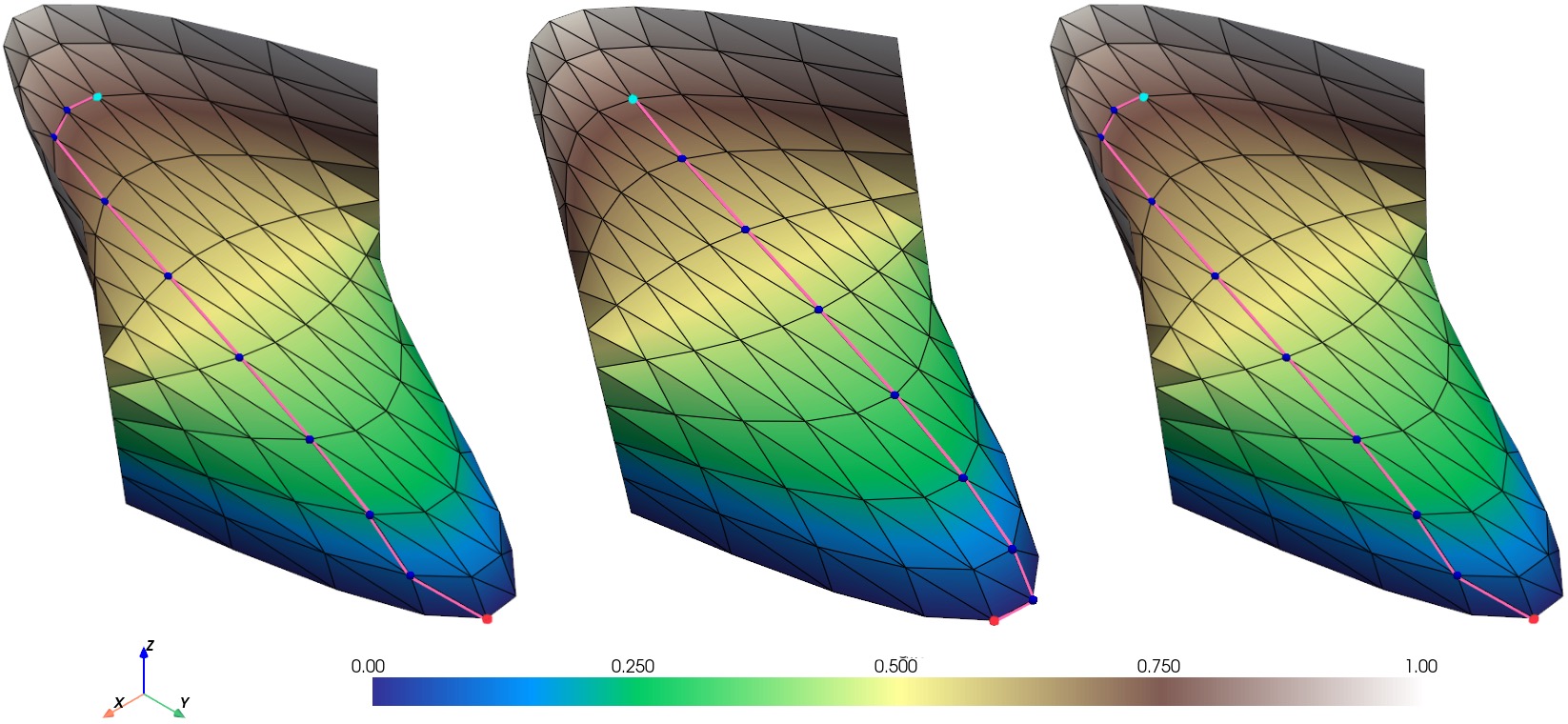}
  
    \put(4,7){\color{red}- Start Point}
    \put(4,9){\color{cyan}- Goal Point}
    \put(4,11){\color{myPink} - Path}
    
    \put(0,51){\color{myBlue} - $\theta_{\text{max}} = \frac{\pi}{3},a_{\text{max}} = 0.9 $}
    \put(0,47){\color{myBlue} - $ v_{\text{max}} = 0.5$}

    \put(34,51){\color{myBlue} - $\theta_{\text{max}} = \frac{\pi}{2}, a_{\text{max}} = 0.9$}
    \put(34,47){\color{myBlue} - $v_{\text{max}} = 0.5$}

    \put(67,51){\color{myBlue} - $\theta_{\text{max}} = \frac{\pi}{3}, a_{\text{max}} = 0.5$}
    \put(67,47){\color{myBlue} - $ v_{\text{max}} = 0.5$}
    
     \put(46,4){ Elevation}
  \end{overpic}
  \caption{An illustration using Mesh 2 and an identical start and goal scenarios with varying acceleration and curvature constraints, demonstrating how the MIKD Planner adapts the generated path to satisfy the constraints.}
    \label{fig:mesh2}
\end{figure*}
The triangular mesh is a graph \( G = (N, E) \), where \( N \) represents the set of vertices (or nodes), and \( E \) denotes the set of edges. The objective is to determine a minimal time path from a given start vertex \( n_1 \in N \) to a goal vertex \( n_f \in N \), subject to constraints on initial and final velocities, maximum velocity, maximum acceleration between any two points along the path, and the curvature.
\subsection{Problem Definitions}
The minimal time path problem is  formulated as:
\begin{equation}
    \text{Minimize: }  T = \mathbf{x} \cdot \mathbf{t} = \sum_{i} x_i \cdot t_i,
    \label{eq:objective}
\end{equation}
Subject to the constraints:
\begin{equation}
    \sum x_{n_a , n_k} =\sum x_{n_k , n_b}, \quad \forall n_k \in N \setminus \{n_1, n_f\}.
    \label{eq:flow}
\end{equation}
\begin{equation}
    - \sum x_{n_a , n_1} + \sum x_{n_1 , n_b} = 1.
        \label{eq:flow_start}
\end{equation}
\begin{equation}
    - \sum x_{n_a , n_f} + \sum x_{n_f , n_b} = -1.
        \label{eq:flow_target}
\end{equation}
\begin{equation}
    x_k + x_l \leq 1 + z_{kl}.
    \label{eq:curvature}
\end{equation}
\begin{equation}
    2 v_i = v_{n_a} + v_{n_b},
    \label{eq:vel1}
\end{equation}
\begin{equation}
    v_{n_0} = \kappa, \quad v_{n_f} = \gamma,
    \label{eq:vel2}
\end{equation}
\begin{equation}
    v_i \le v_{max} \quad \forall i
        \label{eq:vel3}
\end{equation}
\begin{equation}
    v_i \geq 0,
    \label{eq:nonneg_velocity}
\end{equation}
\begin{equation}
          a_i \leq a_{max}
             \label{eq:acc1}
\end{equation}
where $T$ is the total time to minimise, representing the travel time along the path. $\mathbf{x}$ is a vector of binary decision variables $x_i$, where $x_i = 1$ if edge $i$ is included in the path, and $x_i = 0$ otherwise. $x_{n_a, n_k}$ is a binary variable indicating whether the edge from node $n_a$ to node $n_k$ is part of the path ($x_{n_a, n_k} = 1$) or not ($x_{n_a, n_k} = 0$).
$\mathbf{t}$ is a vector of travel times $t_i$, where $t_i$ is the time to traverse edge $i$. $t_i$ is the time required to traverse edge $i$ in the path. $N$ is the set of all nodes in the network, with $n_1$ as the starting node and $n_f$ as the target (final) node. $n_k$ is an intermediate node in the network, where $n_k \in N \setminus \{n_1, n_f\}$ excludes the start and end nodes. $n_a, n_b$ are nodes representing the start and end of an edge, respectively, in the path. $z_{kl}$ is a binary-valued function that indicates whether the curvature constraint between consecutive edges $k$ and $l$ is satisfied. $v_i$ is the velocity associated with edge $i$ in the path. $v_{n_a}, v_{n_b}$ are the velocities at nodes $n_a$ and $n_b$, respectively. $v_{n_0}$ is the initial velocity at the starting node, set to a constant $\kappa$. $v_{n_f}$ is the final velocity at the target node, set to a constant $\gamma$. $v_{max}$ is the maximum allowable velocity for the vehicle. $a_i$ is the acceleration along edge $i$. $a_{max}$ is the maximum allowable acceleration for the vehicle, where $ a_{max} > 0 $ to ensure well-posedness.
In this work, a subscripted variable (e.g., $x_i$, $e_i$, $a_i$, $v_i$) represents an edge $i$. When the edge's direction matters, the notation $\cdot_{n_a,n_b}$ (e.g., $x_{n_a,n_b}$, $e_{n_a,n_b}$) indicates a directed edge from node $n_a$ to node $n_b$.
This solution determines the minimal time path while respecting vehicle acceleration and steering angle limits, adhering to the velocity profile, and the necessary acceleration controls for each edge.

\subsection{Objective Function}

The total time $T$ to traverse the path, expressed as \eqref{eq:objective}, minimising this yields the minimal time path.

Each edge \( e_{n_a, n_b} \in E \), linking nodes \( n_a \) and \( n_b \), is associated with a binary decision variable \( x_{n_a, n_b} \) that indicates whether the edge is part of the path:
\begin{equation}
    x_{n_a, n_b} =
    \begin{cases}
        1 & \text{if edge } e_{n_a, n_b} \text{ is included in the path,} \\
        0 & \text{otherwise.}
    \end{cases}
\end{equation}
The binary variables \( x_{n_a, n_b} \) are stacked into a vector \( \mathbf{x} \):
\begin{equation}
    \mathbf{x} = 
    \begin{pmatrix}
        x_1 \\
        x_2 \\
        \vdots \\
        x_{2m}
    \end{pmatrix},
\end{equation}
where \( m \) is the total number of mesh edges, each mesh edge counts bidirectionally. Here, \( x_i = 1 \) if edge \( i \) is in the path, and \( x_i = 0 \) otherwise.

 The vector of average velocities \( \mathbf{v} \), where \( v_i \) is the average velocity for edge \( i \). For an edge \( e_i = (n_a, n_b) \) from node \( n_a \) to \( n_b \) with average velocity \( v_i \) and distance \( d_i \), the velocities at \( n_a \) and \( n_b \) are \( v_{n_a} \) and \( v_{n_b} \), respectively, satisfying \eqref{eq:vel1}

The vector of edge lengths is \( \mathbf{d} \) , where \( d_i \) is the known length of edge \( i \). The time to traverse edge \( i \) is given by:
\begin{equation}
    t_i = \frac{d_i}{v_i},
\end{equation}
constrained by \eqref{eq:nonneg_velocity} to ensure time remains non-negative.

\subsection{Constraints}
The constraints ensure path continuity, regulate the velocity profile, and enable smooth transitions between path segments. These constraints are essential for maintaining physical feasibility.

\subsubsection{Non-Terminal Vertices}
For each non-terminal vertex \(n_k \in N \setminus \{n_1, n_f\}\),  the sum of incoming edges \(x_{n_a , n_k}\) from some adjacent node $n_a$ to $n_k$ must equal the sum of outgoing edges \(x_{n_k , n_b}\) from $n_k$ to some adjacent node $n_b$ \eqref{eq:flow}: This constraint ensures that the path is continuous.
\subsubsection{Start Vertex Constraint}
At the start vertex \(v_1\), the net flow is constrained to 1 \eqref{eq:flow_start}:
This constraint ensures that the path starts at the source node and there are no loops.
\subsubsection{Target Vertex Constraint}
At the target vertex \(v_n\), the net flow is constrained to \(-1\) \eqref{eq:flow_target}:
This constraint ensures that the path ends at the target node and that there are no loops.
\subsubsection{Kinodynamic Constraints}
To generate feasible trajectories for a car-like vehicle, the path must satisfy motion-related constraints, namely curvature, velocity, and acceleration limits. These constraints reflect the vehicle’s physical capabilities and influence the geometry and timing of the trajectory.

\setlength{\parindent}{0pt}
\textbf{Curvature Constraint :} 
        Limiting the path's curvature enforces the vehicle's steering angle constraint.
    
        Given an edge vector  
        \begin{equation}
            \mathbf{e}_k = (e_{k,x},\, e_{k,y},\, e_{k,z}),
        \end{equation}
        define its xy-projection as  
        \begin{equation}
            \mathbf{e}_k^{xy} = (e_{k,x},\, e_{k,y},\, 0),
        \end{equation}
        which captures the heading direction.  
        
        The yaw angle between two consecutive edges, \( \mathbf{e}_k \) and \( \mathbf{e}_l \), is computed using their xy-projections: 
        \begin{equation}
            \cos(\theta_{kl}^{\text{yaw}}) = \frac{\mathbf{e}_k^{xy} \cdot \mathbf{e}_l^{xy}}{\|\mathbf{e}_k^{xy}\| \, \|\mathbf{e}_l^{xy}\|}.
        \end{equation}
        Define a yaw threshold \( \theta_{\text{max,yaw}} \) and its cosine \( \alpha_{\text{yaw}} = \cos(\theta_{\text{max,yaw}}) \). The yaw constraint is then enforced as follows:  
        \begin{equation}
            z_{kl}^{\text{yaw}} =
            \begin{cases}
                1, & \text{if } \cos(\theta_{kl}^{\text{yaw}}) > \alpha_{\text{yaw}}, \\
                0, & \text{otherwise}.
            \end{cases}
        \end{equation}
        
        Furthermore, the pitch angle of an edge \( \mathbf{e}_k \) is defined as the signed angle between the edge and its xy-projection:  
        \begin{equation}
            \phi_k = \operatorname{atan2}(e_{k,z}, \|\mathbf{e}_k^{xy}\|),
        \end{equation}
        where \( \phi_k > 0 \) for upward slopes and \( \phi_k < 0 \) for downward slopes.
        
        To ensure the vehicle can navigate the path, we impose two pitch-related constraints: 
        
        1. \textbf{Absolute Pitch Limit:} Each edge must not be too steep:  
        \begin{equation}
            |\phi_k| < \phi_{\text{max}},
        \end{equation}
        where \( \phi_{\text{max}} \) is the maximum allowable pitch angle.  
        
        2. \textbf{Pitch Transition Limit:} The change in pitch between consecutive edges must be limited to prevent sharp vertical transitions:  
        \begin{equation}
            |\phi_l - \phi_k| < \theta_{\text{max,pitch}},
        \end{equation}
        where \( \theta_{\text{max,pitch}} \) is the maximum allowable pitch transition.
        
        Thus, the pitch constraint for the transition is:  
       \begin{equation}
z_{kl}^{\text{pitch}} = 
\begin{cases}
1, & \text{if } |\phi_k|, |\phi_l| < \phi_{\text{max}} \text{ and } |\phi_l - \phi_k| < \theta_{\text{max,pitch}}, \\
0, & \text{otherwise}
\end{cases}
\end{equation}
        
        Finally, a transition between edges is allowed only if both the yaw and pitch constraints are satisfied:  
        \begin{equation}
            z_{kl} =
            \begin{cases}
                1, & \text{if } z_{kl}^{\text{yaw}} = 1 \text{ and } z_{kl}^{\text{pitch}} = 1, \\
                0, & \text{otherwise}.
            \end{cases}
        \end{equation}
        Therefore, a transition is valid only if the change in heading does not exceed \( \theta_{\text{max,yaw}} \), the pitch of each edge is within \( \phi_{\text{max}} \), and the difference in pitch between consecutive edges is less than \( \theta_{\text{max,pitch}} \).
        
        For consecutive edges \( x_k, x_l \) and \( z_{kl} \), the  constraint \eqref{eq:curvature} is applied.        
        This constraint ensures that if the angle between \( x_k \) and \( x_l \) exceeds the threshold, then only one of the edges is used. The function $z_{kl}$ is precomputed for the specific mesh environment
        
\textbf{Velocity Constraints :}
            For start and end nodes \(n_0,  n_f \in N\), the velocity \(v\) must satisfy:
            where \(\kappa\) and \(\gamma\) are the initial and final velocities, respectively \eqref{eq:vel2}.
            Also, the edge velocity \(v_i\) along the path \(i\) must stay below the maximum velocity \eqref{eq:vel3}:
            Which constrains the velocity profile of the generated path.
          
\textbf{Acceleration Constraints: }
            Assuming a constant acceleration along each edge, the acceleration along $e_{n_a, n_b}$, derived from the kinematic equation as,
            \begin{equation}
                a_{n_a, n_b} = \frac{{v_{n_b}}^2 - {v_{n_a}}^2}{2d_i},
            \end{equation}           
 Since for every edge $e_{n_a, n_b}$ there is a corresponding twin edge $e_{n_b, n_a}$, with acceleration $a_{n_b, n_a} = - a_{n_a, n_b}$, the constraint \eqref{eq:acc1} also implies a lower bound $-a_{max}$ for an arbitrary edge $e_i$. Stated informally, for an edge to violate the lower bound $-a_{max}$, the twin of this edge would have to violate the upper bound, $a_{max}$, and since this twin is constrained by \eqref{eq:acc1}, the lower bound is enforced implicitly. 

\textbf{Proposition 1:} If the constraint $a_{i} \leq a_{\max}$ holds for all edges $e_{i} \in E$, then $a_{i} \geq -a_{\max}$ also holds for all edges.

\textbf{Proof by contradiction:}

Assume there exists an edge $e_{n_a, n_b} \in E$ such that $a_{n_a, n_b} < -a_{\max}$.

By the twin edge property, there exists $e_{n_b, n_a} \in E$ with:
\begin{align}
a_{n_b, n_a} &= \frac{v_{n_a}^2 - v_{n_b}^2}{2d_{n_b, n_a}} \\
&= -\frac{v_{n_b}^2 - v_{n_a}^2}{2d_{n_a, n_b}} \\
&= -a_{n_a, n_b}
\end{align}
(since $d_{n_b, n_a} = d_{n_a, n_b} = d_i$).

$\therefore$
\begin{align}
a_{n_b, n_a} &= -a_{n_a, n_b} \\
&> -(-a_{\max}) \\
&= a_{\max}
\end{align}

This contradicts the given constraint that $a_{i} \leq a_{\max}$.

Hence, no edge can satisfy $a_{n_a, n_b} < -a_{\max}$, which means $a_{n_a, n_b} \geq -a_{\max}$ for all edges.

\textbf{Conclusion:} The constraint $a_i \leq a_{\max}$ for all edges implicitly enforces $-a_{\max} \leq a_i \leq a_{\max}$. $\square$

\subsection{Relaxation and Solution}
Recalling the objective defined in Equation~\eqref{eq:objective}, we now express it in terms of $\mathbf{d}$ and $\mathbf{v}$ as:
\begin{equation}
    T = \mathbf{x} \cdot \frac{\mathbf{d}}{\mathbf{v}}.
\end{equation}
 Let \begin{equation}
    \mathbf{s} = \frac{1}{\mathbf{v}}, \quad \text{so} \quad \mathbf{s} \mathbf{v} = 1,
    \label{eqn:var_change}
\end{equation}
then the objective function transforms into:  
\begin{equation}
    \mathbf{x} \cdot (\mathbf{d} \cdot \mathbf{s})
\end{equation}
which is now bilinear in \( \mathbf{x} \) and \( \mathbf{s} \). This allows the use of McCormick envelopes \cite{mccormick1976computability} to handle the bilinear term \( \mathbf{x} \cdot \mathbf{s} \).  

\subsection{McCormick Envelopes}

\begin{figure}[htp]
  \centering
  \begin{overpic}[width=\linewidth]{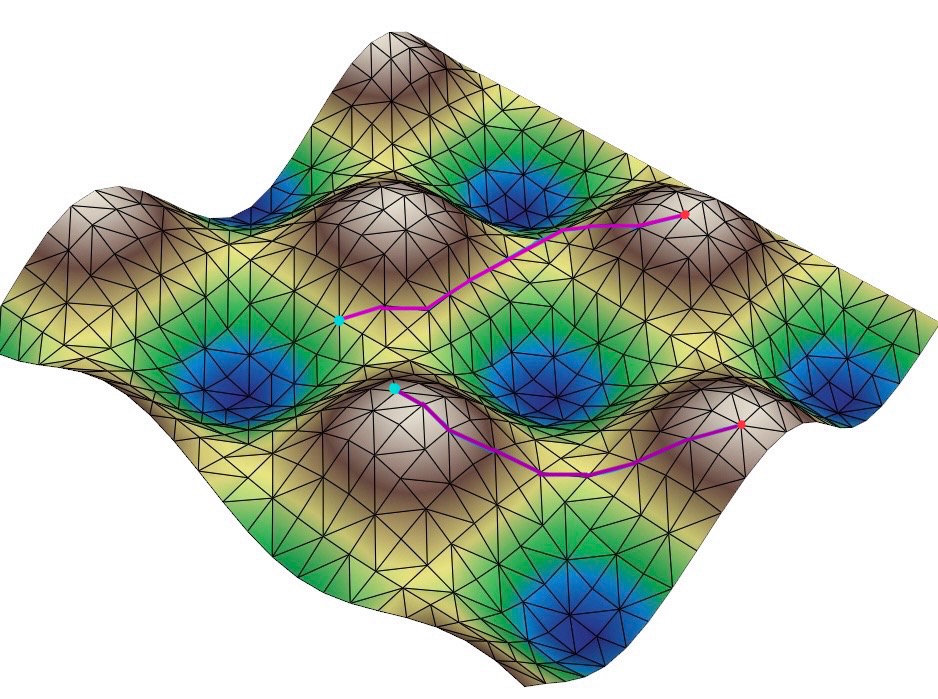}
  
    \put(4,7){\color{red}- Start Point}
    \put(4,11){\color{cyan}- Goal Point}
    \put(4,15){\color{purple} - Path}
       
  \end{overpic}
  \caption{A simulation result using Mesh 1 demonstrates how the MIKD Planner generates paths avoiding potholes and steep descents.}
    \label{fig:mesh1}
\end{figure}
\begin{figure}
    \centering
    \includegraphics[width=\linewidth]{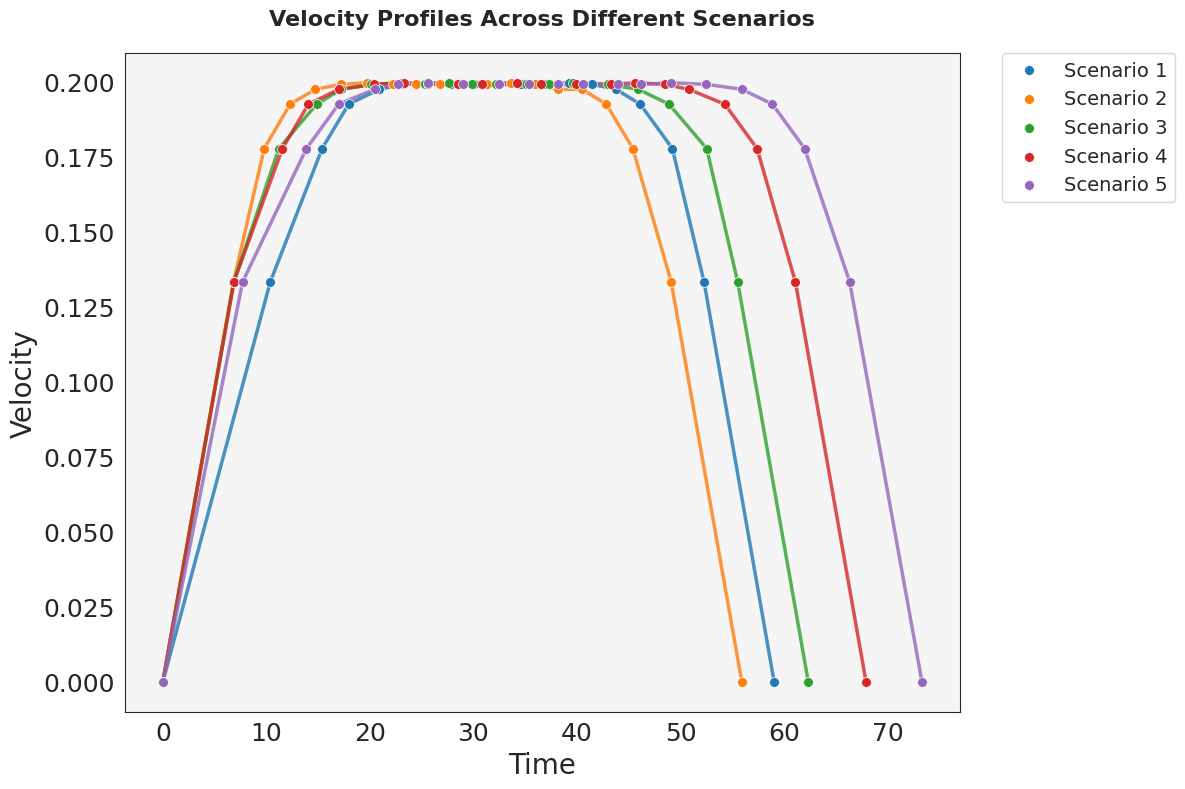}
    \caption{Velocity profiles for five scenarios as a function of time. MIKD demonstrates a continuous velocity transition from initial to final values, adhering to acceleration constraints.($\theta_{\text{max}} = \frac{\pi}{3},a_{\text{max}} = 0.5, v_{\text{max}} = 0.2$)
    }
    \label{fig:v_profiles}
\end{figure}
\begin{figure}
    \centering
    \includegraphics[width=\linewidth]{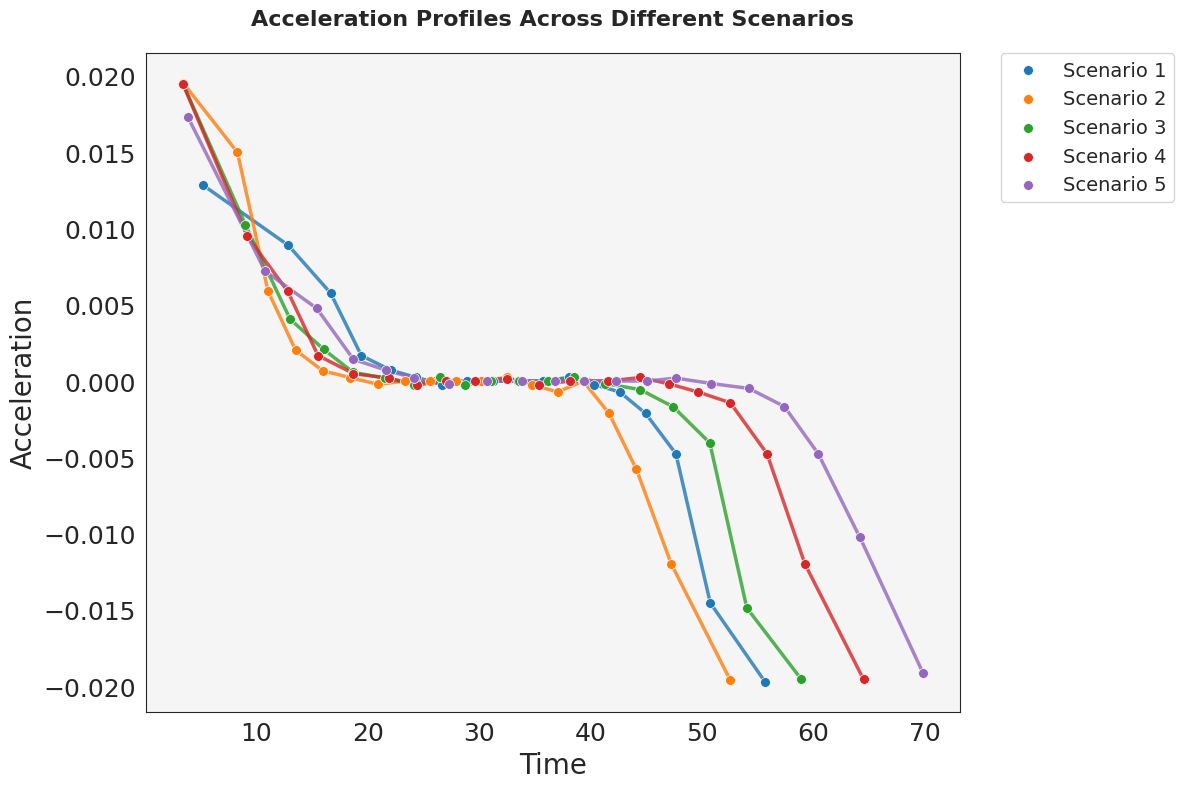}
    \caption{Acceleration profiles for five scenarios as a function of time. MIKD ensures a smooth acceleration transition from initial to final values, minimising jerk while adhering to dynamic constraints. ($\theta_{\text{max}} = \frac{\pi}{3},a_{\text{max}} = 0.5, v_{\text{max}} = 0.2$)}
    \label{fig:acc_profiles}
\end{figure}
\begin{table}[htbp]
\centering
\caption{Time and Velocity Data Across Different Scenarios ($\theta_{\text{max}} = \frac{\pi}{3},a_{\text{max}} = 0.5, v_{\text{max}} = 0.2$)}
\label{table:time_velocity}
\resizebox{\linewidth}{!}{%
\begin{tabular}{c c| c c| c c| c c| c c c}
\hline
\multicolumn{2}{c|}{\textbf{Scenario 1}} & \multicolumn{2}{c|}{\textbf{Scenario 2}} & \multicolumn{2}{c|}{\textbf{Scenario 3}} & \multicolumn{2}{c|}{\textbf{Scenario 4}} & \multicolumn{2}{c}{\textbf{Scenario 5}} \\
\hline
\textbf{Time (s)} & \textbf{Velocity (m/s)} & \textbf{Time (s)} & \textbf{Velocity (m/s)} & \textbf{Time (s)} & \textbf{Velocity (m/s)} & \textbf{Time (s)} & \textbf{Velocity (m/s)} & \textbf{Time (s)} & \textbf{Velocity (m/s)} \\
\hline
0.000 & 0.000000 & 0.000 & 0.000000 & 0.000 & 0.000000 & 0.000 & 0.000000 & 0.000 & 0.000000 \\
10.377 & 0.133333 & 6.822 & 0.133333 & 6.854 & 0.133333 & 6.860 & 0.133333 & 7.702 & 0.133333 \\
15.377 & 0.177778 & 9.788 & 0.177778 & 11.203 & 0.177778 & 11.549 & 0.177778 & 13.867 & 0.177778 \\
17.945 & 0.192593 & 12.299 & 0.192593 & 14.856 & 0.192593 & 14.060 & 0.192593 & 16.980 & 0.192593 \\
20.901 & 0.197531 & 14.737 & 0.197531 & 17.240 & 0.197531 & 17.016 & 0.197531 & 20.448 & 0.197531 \\
23.209 & 0.199177 & 17.191 & 0.199177 & 20.147 & 0.199177 & 20.426 & 0.199177 & 22.740 & 0.199177 \\
25.499 & 0.199726 & 19.682 & 0.199726 & 23.038 & 0.199726 & 23.317 & 0.199726 & 25.631 & 0.199726 \\
27.783 & 0.199177 & 22.202 & 0.199177 & 25.318 & 0.199177 & 25.597 & 0.199177 & 29.022 & 0.199177 \\
30.066 & 0.199177 & 24.490 & 0.199177 & 27.598 & 0.199726 & 28.492 & 0.199177 & 32.418 & 0.199177 \\
32.349 & 0.199177 & 26.772 & 0.199177 & 29.878 & 0.199177 & 30.774 & 0.199177 & 35.313 & 0.199177 \\
34.631 & 0.199177 & 29.054 & 0.199177 & 32.160 & 0.199177 & 34.165 & 0.199726 & 38.208 & 0.199177 \\
36.913 & 0.199177 & 31.336 & 0.199177 & 35.055 & 0.199177 & 36.497 & 0.199177 & 40.543 & 0.199177 \\
39.193 & 0.199726 & 33.616 & 0.199726 & 37.337 & 0.199177 & 39.893 & 0.199177 & 43.939 & 0.199177 \\
41.473 & 0.199177 & 35.896 & 0.199177 & 39.617 & 0.199726 & 43.289 & 0.199177 & 46.221 & 0.199177 \\
43.765 & 0.197531 & 38.188 & 0.197531 & 43.008 & 0.199177 & 45.621 & 0.199726 & 49.112 & 0.199726 \\
46.096 & 0.192593 & 40.489 & 0.197531 & 45.915 & 0.197531 & 48.512 & 0.199177 & 52.503 & 0.199177 \\
49.209 & 0.177778 & 42.820 & 0.192593 & 48.871 & 0.192593 & 50.804 & 0.197531 & 55.913 & 0.197531 \\
52.267 & 0.133333 & 45.388 & 0.177778 & 52.524 & 0.177778 & 54.272 & 0.192593 & 58.869 & 0.192593 \\
59.034 & 0.000000 & 49.094 & 0.133333 & 55.514 & 0.133333 & 57.385 & 0.177778 & 61.982 & 0.177778 \\
 &  & 55.912 & 0.000000 & 62.333 & 0.000000 & 61.091 & 0.133333 & 66.331 & 0.133333 \\
 &  &  &  &  &  & 67.910 & 0.000000 & 73.308 & 0.000000 \\
\hline
\end{tabular}%
}
\end{table}
The McCormick Envelope is a convex relaxation technique for optimising bilinear non-linear programming problems. It replaces each bilinear term with a new variable and adds four constraints, converting the problem into a solvable convex linear program. This method assumes known minimum and maximum values to construct convex and concave envelopes, thereby simplifying analysis and solution.
We use this to relax the original problem, simplifying its analysis and enabling a more tractable solution.

If \( v \) is bounded by \( v_{\min} \leq v \leq v_{\max} \), then \( s \) is also bounded as:  

\begin{equation}
 \frac{1}{v_{\max}} \leq s \leq \frac{1}{v_{\min}}   
\end{equation}

Now, let \( \mathbf{h} = \mathbf{x} \cdot \mathbf{s} \), applying McCormick relaxation:  
\begin{itemize}
    \item Convex Underestimators (Lower Bounds):
    \begin{equation}
        h \geq x_L s + x s_L - x_L s_L
        \label{eqn:mc01}
    \end{equation}
    \begin{equation}
        h \geq x_U s + x s_U - x_U s_U
        \label{eqn:mc02}
    \end{equation}
    \item Concave Overestimators (Upper Bounds)
    \begin{equation}
        h \leq x_U s + x s_L - x_U s_L
        \label{eqn:mc03}
    \end{equation}
    \begin{equation}
        h \leq x_L s + x s_U - x_L s_U
        \label{eqn:mc04}
    \end{equation}
\end{itemize} 
  
where \( x_L = 0 \), \( x_U = 1 \), and \( s_L = \frac{1}{v_{\max}} \), \( s_U = \frac{1}{v_{\min}} \).

\vspace{0.5cm}
This substitution simplifies \eqref{eqn:mc01}, \eqref{eqn:mc02}, \eqref{eqn:mc03} and \eqref{eqn:mc04} to :
 \begin{gather}
    x s_L \leq h \\
    h \leq x s_U  \\
    s + x s_U - s_U \leq h  \\
    h \leq s + x s_L - s_L 
\end{gather}

If \(v_{min} = 0\), then a large positive value can be set for $s_U$.

The objective becomes:
\begin{equation}
    T = \min_{\mathbf{h}} \mathbf{h} \cdot \mathbf{d}
    \label{eqn:relaxed_obj}
\end{equation}

This converts the original objective into a mixed-integer linear programming (MILP) objective. 

However, \eqref{eqn:var_change} introduces a bilinear constraint, this can be relaxed similarly using McCormick envelopes as:

\begin{gather}
    1 \geq s_L v + s v_{\min} - s_L v_{\min},\label{eq:mc5} \\
    1 \geq s_U v + s v_{\max} - s_U v_{\max},\label{eq:mc6} \\
    1 \leq s_U v + s v_{\min} - s_U v_{\min},\label{eq:mc7} \\
    1 \leq s_L v + s v_{\max} - s_L v_{\max} \label{eq:mc8}.
\end{gather}

Equations \eqref{eq:mc5}, \eqref{eq:mc6},\eqref{eq:mc7},and \eqref{eq:mc8} can then serve as convex relaxations for \eqref{eqn:var_change}.
The acceleration constraint \eqref{eq:acc1}, can be rewritten as:
\begin{equation}
    (v_{n_b} - v_{n_a})(v_{n_b} + v_{n_a}) \leq 2d_i a_{\max}
    \label{eqn:rewrite_acc}
\end{equation}
Let \begin{gather}
    \mu = v_{n_b} - v_{n_a} ,\quad \rho = v_{n_b} + v_{n_a}
\end{gather}
The constraint \eqref{eqn:rewrite_acc}, can be written as
\begin{equation}
    \mu \rho \leq 2d_i a_{\max}
\end{equation}\text{Since, }
\begin{gather}v_{\min} \leq v_{n_a}, v_{n_b} \leq v_{\max} \\
-(v_{\max} - v_{\min}) \leq \mu \leq v_{\max} - v_{\min} \\
2v_{\min} \leq \rho \leq 2v_{\max} \\
\mu_{\max} = v_{\max} - v_{\min}, \quad \mu_{\min} = -(v_{\max} - v_{\min}) \\
\rho_{\max} = 2v_{\max}, \quad \rho_{\min} = 2v_{\min}
\end{gather}

And once again applying McCormick relaxation, let 
\begin{equation}
    \lambda = \mu\rho 
    \label{eqn:lambda}
\end{equation}
such that:
\begin{equation}
    \lambda \leq 2d_i a_{\max}
\end{equation}
The equation \eqref{eqn:lambda} relaxes to:
\begin{gather}
\lambda \ge \mu_{\min}\,\rho + \mu\,\rho_{\min} - \mu_{\min}\,\rho_{\min},\\[1mm]
\lambda \ge \mu_{\max}\,\rho + \mu\,\rho_{\max} - \mu_{\max}\,\rho_{\max},\\[1mm]
\lambda \le \mu_{\max}\,\rho + \mu\,\rho_{\min} - \mu_{\max}\,\rho_{\min},\\[1mm]
\lambda \le \mu_{\min}\,\rho + \mu\,\rho_{\max} - \mu_{\min}\,\rho_{\max}.
\end{gather}

Thus, the original problem is solvable in a relaxed form as a minimisation of \eqref{eqn:relaxed_obj}, subject to the constraints mentioned above.
This relaxed problem can be solved efficiently using an optimisation solver. Figures \ref{fig:v_profiles},  \ref{fig:acc_profiles} and Table \ref{table:time_velocity} show that the solution generates smooth velocity and acceleration profiles.

The scenarios shown in Figures \ref{fig:v_profiles},  \ref{fig:acc_profiles} and Table \ref{table:time_velocity} use the same start and end positions as listed in Table \ref{table:evaluation_scenarios}, but the constraints are not from Table \ref{table:evaluation_constraints}. Both scenarios and constraints are selected without loss of generality to highlight the characteristic velocity patterns.

\section{Simulated Evaluation}
There is no published implementation of a global planner for car-like vehicles on a 3D mesh that incorporates kinodynamic constraints.
To study the performance of the proposed algorithm, we have developed an easy-to-use software package for simulating 3D mesh planning.
Afterwards, we compared the proposed global planner with two state-of-the-art sampling-based planners: MPPI and log-MPPI \cite{9834098}. 

The cost function \eqref{eq:cost} was used in the MPPI and log-MPPI implementations, aiming to achieve the same effects as the MIKD objective and constraints, in minimising acceleration and steering angle, as well as path length.

\begin{equation}
 \label{eq:cost}
\begin{aligned}
J = &\ \|\mathbf{p}_{\text{next}} - \mathbf{p}_{\text{target}} \| \\
&+ \left( \max\left(0, |u_0| - a_{\max} \right) / a_{\max} \right)^2 \\
&+ \left( \max\left(0, |u_1| - \dot{\delta}_{\max} \right) / \dot{\delta}_{\max} \right)^2.
\end{aligned}
\end{equation}
\( \mathbf{p}_{\text{next}} \) and \( \mathbf{p}_{\text{target}} \) are the 3D position vectors of the next position and target, respectively, it provides the planner with a sense of the globally shortest distance to the target.\( u_0 \) represents the acceleration control input.\( u_1 \) represents the steering rate control input.
\( a_{\max} \) and \( \dot{\delta}_{\max} \) are the maximum acceleration and steering rate, respectively.

Equal weights are used in this cost, chosen based on empirical tuning and observed convergence behaviour. The quadratic penalty structure for constraint violations provides inherent scaling when values exceed their limits, effectively balancing the influence of control penalties against the spatial distance term. In practice, introducing unequal weights often led to worse performance, and in some scenarios, logMPPI failed to converge.

\subsection{Evaluation Metrics}
The selection of these evaluation metrics is grounded in the shared objective across all three algorithms: to compute the shortest path while respecting both acceleration and steering constraints. Specifically, the metrics were chosen to capture the trade-off involved in optimising path length, constraint adherence, and minimising execution time.

\begin{itemize} 
     \item \textbf{Constraint Error $\Pi$}: The path curvature, velocity, and acceleration must remain within predefined constraints; the constraint error quantifies deviations from these constraints, denoted by $\Pi$. This metric measures the extent of violation:
   \begin{equation}
    \begin{aligned}
        \Pi &= \sum_{i=1}^{m-2} \max(0,| \theta_{i,i+1} | - \theta_{max}) \\
          &\quad + \sum_{i=1}^{m-1} \max(0, |a_i| - a_{\max}) \\ 
          &\quad + \sum_{i=1}^{m} \max(0, v_i - v_{\max}),
    \end{aligned}
  \end{equation}
    a smaller value represents a better performance.
    \item \textbf{Path Length $\Delta$}:  
    The path length measures the total distance travelled from start to goal, normalised by the direct Euclidean distance. It is computed as the sum of the Euclidean distances between consecutive waypoints, subtracting the direct Euclidean distance from start to goal, and then dividing by this direct distance:

   \begin{equation}
       \Delta = \frac{( \sum_{i=1}^{m-1} \left\| p_{i+1}  - p_i \right\| ) - \left\| p_1 - p_m \right\|}{\left\| p_1 - p_m \right\|}
   \end{equation}

    Where:
    - \( p_i \) and \( p_{i+1} \) are consecutive positions,
    - \( m \) is the number of waypoints,
    - \( \left\| p_1 - p_m \right\| \) is the direct Euclidean distance from start to goal.
    
    \item \textbf{Execution Time $\tau$}: This measures the time in seconds for the algorithm to generate a path.
 
\end{itemize}
The chosen evaluation metrics, Constraint Error ($\Pi$), Path Length ($\Delta$), and Execution Time ($\tau$), provide a holistic assessment of algorithm performance. $\Pi$ quantifies compliance with dynamic and kinematic constraints, $\Delta$ reflects path efficiency relative to an ideal trajectory, and $\tau$ represents computational cost. Collectively, they offer a balanced measure of path quality and feasibility under realistic motion constraints.

\subsection{Experiment Design}
The experiments study the effectiveness of the planning algorithms in generating paths that lie on the surface of triangulated mesh environments, along with a velocity profile, such that the velocity, acceleration, and turns along the path satisfy user-specified values.
We evaluated the performance of the three planning algorithms in two mesh environments, each with different complexities. The experiments are as follows:
\begin{itemize}
    \item \textbf{Number of Runs}: For each environment, start-goal scenario, and configuration, we ran five independent trials to ensure statistical reliability.
    \item \textbf{Environments}:
    Two triangulated mesh environments evaluate algorithm performance under different geometric complexities:
    \begin{itemize}
        \item \textbf{Mesh 1}: High-complexity, 983 triangular faces, intricate surface with ridges and valleys.
        \item \textbf{Mesh 2}: Low-complexity, 200 triangular faces, smooth surface.
    \end{itemize}
    \item \textbf{Start/Goal Scenarios}: For each environment, we tested five different start-goal pairs, ensuring diverse spatial and topological challenges for the algorithms, as shown in Table \ref{table:evaluation_scenarios}.
    \item \textbf{Constraints Sets}: We used three sets of parameters (\(\theta_{\text{max}}, a_{\text{max}}, v_{\text{max}}\)) to test how varying settings affect performance, summarized in Table \ref{table:evaluation_constraints}. Additionally, the initial and final velocities, \(\kappa\) and \(\gamma\), are set to \(e^{-24}\), which is practically equivalent to zero but avoids zero division errors. 
    Furthermore, the velocity range used in these simulations was selected because of the edge distances of the mesh triangulation, ensuring that the computed time values would be significant.    
\end{itemize}
The experiment consisted of five trials for each unique combination of environment, start-goal pair, and configuration set as summarised in Table \ref{tab:experiment_design}. For valid comparison, we applied min-max normalisation to each dataset column.
\begin{table}[htbp!]
    \caption{Experimental design summarising the key factors and their combinations used to evaluate the planning algorithm across varying environmental complexity, scenario difficulty, and motion constraints.}

    \label{tab:experiment_design}
    \centering
    \begin{tabular}{l|c|p{4cm}}
        \hline
        \textbf{Factor} & \textbf{Counts} & \textbf{Description} \\
        \hline
        \textbf{Environments} & 2 &  Mesh 1 (983 faces), Mesh 2 (200 faces) \\
        \hline
        \textbf{Start-Goal Scenarios} & 5 & Different topological challenges \\
        \hline
        \textbf{Constraint Sets} & 3 & Different constraints (\(\theta_{\text{max}}, a_{\text{max}}, v_{\text{max}}\)) \\
        \hline
        \textbf{Trials} & 5 & Independent runs for statistical reliability \\
        \hline
        \textbf{Total Number of Experiments} & 150 & \(2 \times 5 \times 3 \times 5\) \\
        \hline
    \end{tabular}
\end{table}

\begin{table}[htbp]
\centering
\begin{threeparttable}
\caption{The table shows start and end coordinates, and Euclidean distances, for the evaluation scenarios used in Mesh 1 and Mesh 2. The scenarios test the planning performance under varying spatial and topological configurations across differing environmental complexities.}

\label{table:evaluation_scenarios}
\begin{tabular}{l c c c}
\textbf{Scenario} & \textbf{Start Point (x, y, z)} & \textbf{End Point (x, y, z)} & \textbf{Distance} \\
\hline
\multicolumn{4}{c}{\textbf{Mesh 1}} \\
\hline
1 & (5.00, 9.55, 0.95) & (6.82, 1.36, 0.10) & 8.42 \\
2 & (9.09, 9.55, -0.33) & (7.73, 1.36, 0.20) & 8.31 \\
3 & (5.00, 9.55, 0.95) & (0.91, 0.91, 0.48) & 9.57 \\
4 & (9.09, 10.00, -0.27) & (2.73, 0.91, 0.25) & 11.11 \\
5 & (0.45, 9.55, -0.44) & (8.18, 1.36, 0.19) & 11.27 \\
\hline
\multicolumn{4}{c}{\textbf{Mesh 2}} \\
\hline
6 & (0.56, 0.89, -0.93) & (0.78, 0.22, 0.49) & 1.58 \\
7 & (0.22, 0.89, -0.60) & (0.67, 0.22, 0.66) & 1.50 \\
8 & (1.00, 0.33, 0.00) & (0.11, 0.78, -0.26) & 1.03 \\
9 & (0.33, 1.00, -0.87) & (0.89, 0.11, 0.32) & 1.58 \\
10 & (0.11, 1.00, -0.34) & (0.78, 0.33, 0.32) & 1.15 \\
\hline
\end{tabular}
\begin{tablenotes}
\footnotesize
\item Note: While the geometry processing tools treat mesh units as dimensionless, all coordinates and distances in this study are in meters.
\end{tablenotes}
\end{threeparttable}
\end{table}

\begin{figure}[htbp]
    \centering
    \includegraphics[width=\linewidth]{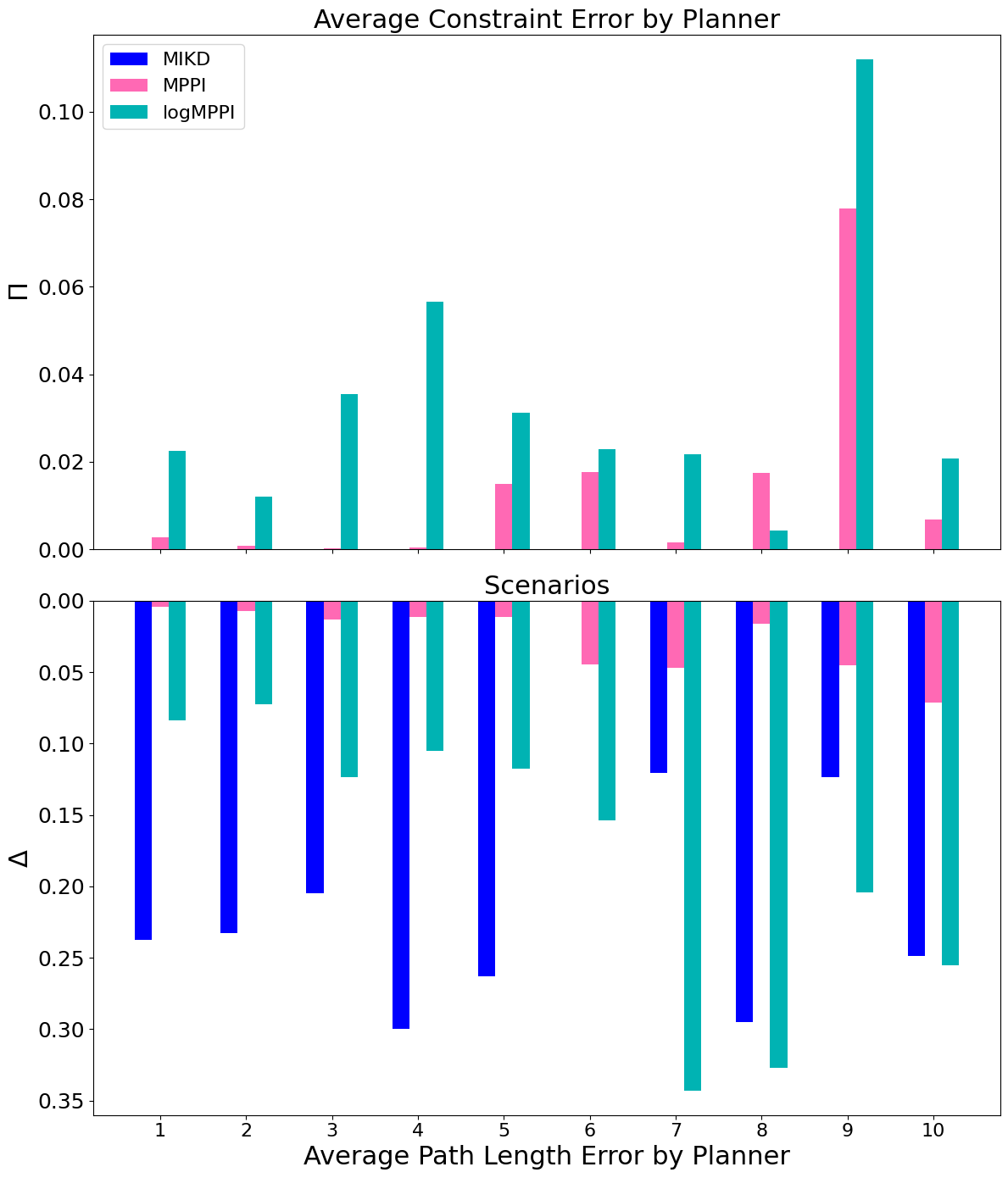}
    \caption{Performance comparison of algorithms based on path length and constraint violations across various scenarios. MIKD consistently outperforms both MPPI and logMPPI in satisfying motion constraints. While MPPI achieves the shortest path length, MIKD's path length is comparable to logMPPI.}
    \label{fig:eval1}
\end{figure}

\begin{figure}[htbp]
    \centering
    \includegraphics[width=\linewidth]{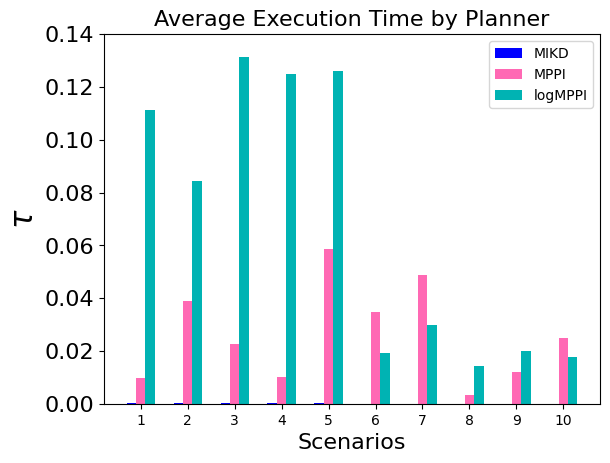}
    \caption{Comparison of execution times for different scenarios across three algorithms: MIKD, MPPI, and logMPPI. Notable differences in time performance are observed, with MIKD consistently showing low values and logMPPI exhibiting greater variability.}
    \label{fig:eval2}
\end{figure}

\begin{table}[htbp]
\centering
\caption{Constraint sets used in evaluation, varying turning angle (\(\theta_{\text{max}}\)), acceleration (\(a_{\text{max}}\)), and velocity (\(v_{\text{max}}\)) to represent different operational limits.}
\label{table:evaluation_constraints}
\begin{tabular}{p{2cm} c c c}
\hline
\textbf{Scenario} & \(\theta_{\text{max}}\) (deg) & \(a_{\text{max}}\) (m/s\(^2\)) & \(v_{\text{max}}\) (m/s) \\
\hline
1 & \( \frac{\pi}{3}\) & 0.50 & 0.90 \\
2 & \( \frac{\pi}{2}\) & 0.90 & 0.50 \\
3 & \( \frac{\pi}{3}\) & 0.50 & 0.50 \\
\hline
\end{tabular}
\end{table}

\begin{table}[htbp]
\centering
\caption{Table of Constraint Errors for Mesh 1 and Mesh 2 Across Multiple Scenarios, Showing MIKD Errors on the Order of \(10^{-7}\), Smaller than MPPI and logMPPI, with Mesh 2 Having Slightly Higher Errors for MPPI and logMPPI}
\label{table:constraint_error}
\resizebox{\linewidth}{!}{%
\begin{tabular}{c|c|c|c|c|c|c|c}
\hline
\multicolumn{4}{c|}{\textbf{Mesh 1}} & \multicolumn{4}{c}{\textbf{Mesh 2}} \\
\hline
\textbf{Scenario} & \textbf{MIKD} & \textbf{MPPI} & \textbf{logMPPI} &
\textbf{Scenario} & \textbf{MIKD} & \textbf{MPPI} & \textbf{logMPPI} \\
\hline
1  & \(3.5243e{-7}\) & 0.00275 & 0.02256 & 6  & \(3.3475e{-7}\) & 0.01770 & 0.02296 \\
2  & \(1.8327e{-7}\) & 0.00078 & 0.01208 & 7  & \(2.9084e{-7}\) & 0.00164 & 0.02178 \\
3  & \(4.6251e{-7}\) & 0.00012 & 0.03555 & 8  & \(1.9072e{-7}\) & 0.01748 & 0.00419 \\
4  & \(2.1476e{-7}\) & 0.00047 & 0.05667 & 9  & \(4.0129e{-7}\) & 0.07798 & 0.11197 \\
5  & \(5.9832e{-7}\) & 0.01498 & 0.03114 & 10 & \(2.5536e{-7}\) & 0.00678 & 0.02070 \\
\hline
\end{tabular}%
}
\end{table}

\begin{table}[htbp]
\centering
\caption{Table of Path Length Errors for Mesh 1 and Mesh 2 Across Multiple Scenarios, Showing MPPI Errors Generally Smaller than MIKD and logMPPI, with logMPPI Exhibiting Larger Errors on Mesh 2.}
\label{table:path_length_error}
\resizebox{\linewidth}{!}{%
\begin{tabular}{c|c|c|c|c|c|c|c}
\hline
\multicolumn{4}{c|}{\textbf{Mesh 1}} & \multicolumn{4}{c}{\textbf{Mesh 2}} \\
\hline
\textbf{Scenario} & \textbf{MIKD} & \textbf{MPPI} & \textbf{logMPPI} &
\textbf{Scenario} & \textbf{MIKD} & \textbf{MPPI} & \textbf{logMPPI} \\
\hline
1  & 0.23749 & 0.00394 & 0.08343 & 6  & 0.00029 & 0.04462 & 0.15378 \\
2  & 0.23254 & 0.00732 & 0.07232 & 7  & 0.12046 & 0.04672 & 0.34302 \\
3  & 0.20462 & 0.01291 & 0.12340 & 8  & 0.29491 & 0.01626 & 0.32738 \\
4  & 0.29973 & 0.01127 & 0.10508 & 9  & 0.12375 & 0.04500 & 0.20438 \\
5  & 0.26316 & 0.01113 & 0.11773 & 10 & 0.24886 & 0.07102 & 0.25510 \\
\hline
\end{tabular}%
}
\end{table}

\subsection{Experimental Setup}
Simulations were run on an ASUS TUF A17 laptop with an AMD Ryzen 7 4800H processor, featuring eight cores, sixteen threads, a 2.90 GHz base frequency, and 8 GB of RAM. We developed a Python-based software for simulating motion planning in a 3D mesh environment and used it to compare the performance of MIKD, MPPI, and log-MPPI. Figures \ref{fig:mesh2} and \ref{fig:mesh1} illustrate some planning results for two different mesh environments, using the software. Furthermore, the proposed global planner is integrated into the ROS/Gazebo Move Base Flex framework, implemented in C++, and optimised using the Gurobi solver \cite{gurobi}.

\begin{table}[htbp]
\centering
\caption{Table of Execution Times for Mesh 1 and Mesh 2 Across Multiple Scenarios, Showing MIKD Times on the Order of \(10^{-4}\) to \(10^{-5}\) Seconds, Faster than MPPI and logMPPI, with Mesh 2 Generally Having Shorter Times for MIKD and logMPPI}
\label{table:execution_time}
\resizebox{\linewidth}{!}{%
\begin{tabular}{c|c|c|c|c|c|c|c}
\hline
\multicolumn{4}{c|}{\textbf{Mesh 1}} & \multicolumn{4}{c}{\textbf{Mesh 2}} \\
\hline
\textbf{Scenario} & \textbf{MIKD} & \textbf{MPPI} & \textbf{logMPPI} &
\textbf{Scenario} & \textbf{MIKD} & \textbf{MPPI} & \textbf{logMPPI} \\
\hline
1  & 3.8166e-04 & 0.00961 & 0.11119 & 6  & 3.2048e-05 & 0.03480 & 0.01936 \\
2  & 3.5229e-04 & 0.03880 & 0.08441 & 7  & 1.7414e-05 & 0.04887 & 0.02970 \\
3  & 3.5702e-04 & 0.02270 & 0.13146 & 8  & 1.7089e-05 & 0.00317 & 0.01439 \\
4  & 3.6392e-04 & 0.01009 & 0.12490 & 9  & 2.6846e-05 & 0.01218 & 0.02006 \\
5  & 3.5243e-04 & 0.05868 & 0.12616 & 10 & 2.4725e-05 & 0.02493 & 0.01763 \\
\hline
\end{tabular}%
}
\end{table}

\subsection{Results and Analysis}
The experimental evaluation shows that the MIKD algorithm performs favourably compared to MPPI and logMPPI in satisfying motion constraints, maintaining reasonable path lengths, and achieving lower execution times. As presented in Tables \ref{table:constraint_error}, \ref{table:path_length_error}, and \ref{table:execution_time}, as well as Figures \ref{fig:eval1} and \ref{fig:eval2}, MIKD demonstrates consistent improvements across various scenarios and mesh environments. These results suggest the potential advantages of MIKD in the tested settings.

MIKD consistently achieves infinitesimal constraint error across all scenarios, while MPPI and logMPPI exhibit varying levels of constraint violations. LogMPPI shows the highest error, reaching 0.11197 in Mesh 2, Scenario 9. These results highlight MIKD's effectiveness in satisfying motion constraints without deviation.
The infinitesimal errors might be due to the relaxation bounds.

While MPPI achieves the shortest path length across all scenarios, this comes at the expense of higher constraint violations. MIKD, on the other hand, maintains a path length comparable to logMPPI while effectively balancing efficiency and constraint satisfaction. Notably, MIKD's path length error is higher than MPPI's, particularly in Mesh 1, indicating it is less efficient at path length. This result suggests that MIKD prioritises constraint adherence without significantly compromising path efficiency.

A key advantage of MIKD is its notably low execution time. As shown in Table \ref{table:execution_time}, MIKD is faster than MPPI and logMPPI.
For instance, in Mesh 1, Scenario 1. This trend holds across all tested scenarios, highlighting MIKD's computational efficiency in constrained environments.

The results indicate that MIKD performs effectively in motion planning for constrained environments. It demonstrates consistent adherence to motion constraints, achieves a balanced trade-off in path length, and exhibits notably shorter execution times compared to MPPI and logMPPI.
These characteristics suggest that MIKD is well-suited for real-time applications, where efficiency and constraint satisfaction are essential.

\subsection{Simulated Robot Experiments}
Furthermore, we performed experiments to verify the effectiveness of the velocity commands generated by our algorithm in moving the robot from a start point to a goal point. The experiments were conducted in a simulated environment using the ROS/Gazebo Move Base Flex framework and a simulation of the Pluto robot and Mesh 1. The robot had to reach predefined goals using the five start-end scenarios, executing five runs of each scenario. Table  \ref{tab:success_rate} summarises the results.

\begin{table}[htbp]
    \centering
    \caption{Success Rate per Scenario}
    \label{tab:success_rate}
    \begin{tabular}{p{2.0cm}|p{1.5cm}|p{1.5cm}|p{2.0cm}}
        \hline
        \textbf{Scenario} & \textbf{MIKD} & \textbf{MPPI} & \textbf{logMPPI} \\
        \hline
        1 & 1.0 & 0.8 & 0.8 \\
        2 & 1.0 & 0.6 & 0.8 \\
        3 & 1.0 & 1.0 & 1.0 \\
        4 & 1.0 & 0.8 & 0.6 \\
        5 & 1.0 & 0.8 & 0.8 \\
        \hline
    \end{tabular}
\end{table}
In the tested scenarios, MIKD optimised the path well, avoiding steep climbs; its edge-based approach sometimes caused unnecessary turns. However, it consistently found a feasible route with suitable acceleration and steering.
MPPI performs well but occasionally leads to potholes or unstable pitch angles. Its sampling-based nature causes performance variations, and the log-MPPI strategy sometimes introduces velocity discontinuities; however, it generally matches the effectiveness of MPPI.

\subsection{Limitations}
The algorithm's accuracy and efficiency depend on the quality of triangulation. Poor triangulation causes suboptimal paths and higher computational costs due to inaccurate terrain representation.

The algorithm is edge-based and may require further refinement using algorithms such as FlipOut \cite{keenan2020}.

Overly lenient velocity constraints, where the maximum velocity value is significantly greater than edge distances, can lead to inefficient paths.

Finer meshes with many edges increase the algorithm's memory use. A divide-and-conquer approach or mesh re-triangulation is needed to manage this.

\section{Discussion}
In this study, we have identified some considerations for terrain triangulation in the context of robotic navigation. The ideal terrain mesh should strike a balance between detail and computational efficiency. A wider triangulation can reduce the number of faces in the mesh, thereby improving computational performance. However, the mesh must be sufficiently fine-grained to capture the relevant terrain features without sacrificing accuracy. 

McCormick envelopes are particularly well-suited to our problem because of the bilinear terms arising in the objective from the coupling of the edge indicator variable $x$ and the edge velocity variable $v$, as well as in the quadratic constraints relating to acceleration. The applicability stems from the fact that the minimum and maximum values of these variables are known. The trade-off is the introduction of several auxiliary variables. Additionally, as shown in Table \ref{table:constraint_error}, the error in the relaxed model's approximation of the solution to the original problem is minimal and negligible in the simulated robot experiments.

To improve the quality of mesh-based optimisation, tightening McCormick relaxation bounds or exploring alternative relaxation techniques could provide more precise results. Additionally, terrain meshes like those used in our work can be efficiently generated through LiDAR scans, and surface reconstruction algorithms such as those proposed by \cite{wiemann2018}.

Our simulations, conducted within the Move Base Flex framework and using the Pluto robot, demonstrate that MIKD is useful for mesh navigation.

Looking forward, future research could investigate the application of smoothing techniques that do not compromise the integrity of the constraints. Moreover, the use of optimisation algorithms which optimise the mesh structure offers promising avenues for further enhancing the performance of mesh-based navigation systems.

\section{Conclusion}
In this work, we presented the Mixed-Integer Kinodynamic (MIKD) Planner, a global planning approach that leverages an explicit formulation of constrained planning as an optimisation problem. MIKD integrates vehicle constraints such as acceleration, velocity, and steering into the planning process to generate feasible paths. Its deterministic nature ensures consistent and reliable performance.
However, MIKD has some limitations: sensitivity to mesh quality, edge-based exploration,  challenges with large-scale meshes, and sensitivity to velocity constraints.
Compared to traditional sampling-based methods, the MIKD planner excels in trajectory feasibility and execution time. By directly optimising kinodynamic constraints, MIKD effectively balances feasibility and efficiency, making it a promising solution for planning in complex, 3D mesh environments.

\section*{Author Contributions: CRediT}

\textbf{Otobong Jerome}: Conceptualisation, Software, Writing – original draft.\\
\textbf{Alexandr Klimchik}: Supervision, Formal analysis, Writing – review and editing, Final approval of the version to be submitted.\\
\textbf{Alexander Maloletov}: Supervision, Formal analysis, Writing – review and editing.\\
\textbf{Geesara Kulathunga}: Supervision, Software, Formal analysis, Writing – review and editing.

\section*{Declaration of Competing Interests}

The authors have no competing interests to declare.

\section*{Funding Sources}

This research did not receive any specific grant from funding agencies in the public, commercial, or not-for-profit sectors.

\section*{Data Availability Statement}
Data will be made available on request.

\bibliographystyle{elsarticle-num} 
\bibliography{ref}

\end{document}